\pdfoutput=1

\documentclass[11pt]{article}

\usepackage[]{ACL2023}

\usepackage{times}
\usepackage{latexsym}

\usepackage[T1]{fontenc}

\usepackage[utf8]{inputenc}

\usepackage{microtype}

\usepackage{inconsolata}

\usepackage{graphics}
\usepackage{graphicx}
\graphicspath{ {./images/} }

\usepackage{tikz}
\usepackage{pgfplots}
\usepackage{colortbl}
\pgfplotsset{compat=1.17}

\usepackage{caption}
\usepackage{subcaption}

\definecolor{xy-axis}{HTML}{FFFDE5}
\definecolor{data-appearance}{HTML}{E5E5FF}
\definecolor{title-subtitle}{HTML}{E5E5E5}
\definecolor{legend}{HTML}{FFE5E5}
\definecolor{annotation}{HTML}{FBE7EB}
\definecolor{grid}{HTML}{F9F2EC}
\definecolor{figure-format}{HTML}{E2F3F8}
\definecolor{others}{HTML}{F2F2F2}

\usepackage{booktabs}
\usepackage{listings}
\usepackage{colortbl}

\title{Do Text-to-Vis Benchmarks Test Real Use of Visualisations?}

\author{Hy Nguyen$^1$$^*$, Xuefei He$^1$, Andrew Reeson$^2$, Cécile Paris$^2$, Josiah Poon$^1$, \\ \and \textbf{Jonathan K. Kummerfeld}$^1$ \\
\begin{tabular}{cc}
  The University of Sydney$^1$ &
  CSIRO's Data61$^2$ \\
\end{tabular}\\
\texttt{nngu0448@uni.sydney.edu.au}$^*$
}

\begin{document}
\maketitle
\begin{abstract}
Large language models are able to generate code for visualisations in response to simple user requests.
This is a useful application and an appealing one for NLP research because plots of data provide grounding for language.
However, there are relatively few benchmarks, and those that exist may not be representative of what users do in practice.
This paper investigates whether benchmarks reflect real-world use through an empirical study comparing benchmark datasets with code from public repositories.
Our findings reveal a substantial gap, with evaluations not testing the same distribution of chart types, attributes, and actions as real-world examples.
One dataset is representative, but requires extensive modification to become a practical end-to-end benchmark. 
This shows that new benchmarks are needed to support the development of systems that truly address users' visualisation needs.
These observations will guide future data creation, highlighting which features hold genuine significance for users. 

\end{abstract}


\section{Introduction}
Text-to-Vis is the task of receiving data and a request for a visualisation expressed in human language and generating code that will produce the visualisation.
A system with this ability would enable faster and more complex data analysis, but there are relatively few benchmark datasets for the task.
Those that do exist either focus on generating a single response~\cite{luo2021synthesizing, srinivasan2021collecting, chen2021plotcoder}, or consider dialogue, but with limited flexibility in code~\cite{shao-nakashole-2020-chartdialogs, song2024marrying}.
Most of these datasets use generated data.
The space of code variation was defined by researchers.
This raises the question of whether these datasets are representative of real-world use of data visualisations.

In this study, we gathered publicly available code from the Stack\footnote{A 6TB collection of open source code from GitHub~\cite{kocetkov2022stack}} to analyse human preferences when making visualisations using libraries across four programming languages: Python, R, Javascript, and Vega.
Since each library has different names for the same visualisation types and properties, we extracted key visualisation code and developed a cross-language mapping for several hundred functions and arguments.\footnote{
For example, a bar plot is produced with \texttt{bar()} and \texttt{barh()} in Python, but \texttt{barplot()} in R.
Our data and code are at \href{https://github.com/giahy2507/text-to-vis-bench-assessment}{https://github.com/giahy2507/text-to-vis-bench-assessment}.
}

Using this aligned data, we analysed user behaviours when making visualisations and identified similarities and differences between real-world and benchmark datasets. 
Our analysis considered the chart types, functions called to define properties, and the arguments that modify how those functions behave. 
We observed that (1) existing benchmarks tend to focus on only one aspect of the Text-to-Vis challenge, either code synthesis, data presentation, or aesthetic attribute adjustment, and (2) only one of the datasets is consistent with real-world use and that dataset is limited by its lack of executability of code outputs.

Success on existing benchmarks does not mean systems are useful for real-world use.
We need new benchmarks that cover all aspects of the problem and are consistent with patterns of use. 
Only then will we be able to measure progress on this valuable and challenging task.

\section{Related Work}
A common approach to creating Text-to-Vis datasets involves automatic synthesis of visualisations followed by human intervention for annotation~\cite{luo2021synthesizing, shao-nakashole-2020-chartdialogs, srinivasan2021collecting, song2024marrying, chen2021plotcoder}. 
Although this approach is straightforward, datasets produced using this method often contain inherent problems.
For instance, nvBench, the largest benchmark dataset for this task, was synthesised from Spider~\cite{yu2018spider}, a text-to-SQL dataset containing several limitations~\cite{suhr-etal-2020-exploring}, and was only partially reviewed by novices and experts for quality assurance, resulting in numerous issues~\cite{li2024visualization}. 
Similarly, ChartDialogs contains limitations as the data for visualisation were automatically generated.
These shortcomings underscore the need for improved methodologies in dataset creation to ensure validity and usability.

Recent research in AI has prioritised ecological validity to enhance benchmark dataset quality across various domains, aiming to align them with real-world applications~\cite{de2020towards, qi2023pragmaticqa, lu2023statcan}. 
Ensuring that the data used to train and test models accurately reflects users' objectives in practical scenarios is crucial.
However, prior research on Text-to-Vis has not considered this aspect of dataset creation.

\section{Data Collection}
\label{sec:real-world-data-collection}
Instead of examining user preferences through chart images or relying on experts to comprehend how visualisations are made, our approach involves the analysis of publicly available programs specifically written for creating visualisations, e.g., line, bar, and scatter charts.
This means we consider a wide range of samples from different programmers and their preferences when making visualisations.
As a result, our analysis can provide a broad understanding of the essential components that are widely used.

We used code from The Stack~\footnote{https://huggingface.co/datasets/bigcode/the-stack-dedup} to conduct our investigation.
We consider four diverse and widely used visualisation libraries:
\textbf{Matplotlib}\footnote{https://matplotlib.org/}, \textbf{Graphics\footnote{https://www.rdocumentation.org/packages/graphics}}, \textbf{ChartJS\footnote{https://www.chartjs.org/docs/latest/}}, and \textbf{Vega-Lite\footnote{https://vega.github.io/vega-lite/}}, which are for Python, R, Javascript, and JSON, respectively.
After downloading, we selected files containing code indicative of visualisation library usage (e.g., \verb|import matplotlib.pyplot as plt|). 
Finally, we used abstract syntax tree (AST) parsers and heuristics to accurately extract library-related variables, function names, arguments, and explicit values.
The details are described in Appendix~\ref{sec:appendix-code-parsing}, while Table~\ref{tab:statistics-of-real-world-data} (upper) presents their statistics.

\begin{table}[t]
\centering
\small
\begin{tabular}{lrrr}
\toprule
           & \# samples &  Proportion  & \# functions \\ \midrule
nb-Matplotlib & 385,338 &  35.89\% & 6,443,220 \\
py-Matplotlib &  464,463 &   3.5\%         &   4,484,368  \\
Graphics   &  6,721   &   17.15\%       &   53,325     \\ 
ChartJS    &  2,714   &   0.0128\%      &   8,847      \\ 
Vega-Lite  &  1,093    &   0.0013\%      &   15,664     \\ \midrule
\textcolor{red}{nvBench}  & 7,241 & & 40,478 \\ 
\textcolor{red}{ChartDialogs}  & 3,284 & & 14,690 \\
\textcolor{red}{PlotCoder}  & 97,706 & & 254,251 \\
\bottomrule
\end{tabular}
\caption{Statistics of real-world and benchmark data. ``Proportion'' indicates the proportion of the library's code in the investigated programming languages. ``nb-Matplotlib'' indicates code from Jupyter notebooks, while ``py-Matplotlib'' indicates code from Python files.}
\label{tab:statistics-of-real-world-data}
\end{table}

We examined three publicly available  benchmark datasets: \textcolor{red}{nvBench}~\cite{luo2021synthesizing},  \textcolor{red}{ChartDialogs}~\cite{shao-nakashole-2020-chartdialogs}, and \textcolor{red}{PlotCoder}~\cite{chen2021plotcoder}.
They vary in settings and scales, as described in Table~\ref{tab:statistics-of-real-world-data} and~\ref{tab:benchmark-description}.
While nvBench and ChartDialogs are end-to-end Text-to-Vis benchmarks, with queries \& data as input and code \& visualisations as output, PlotCoder is purely a code synthesis dataset, with no data or visualisations.
Appendix~\ref{sec:appendix_benchmark_examples} shows examples from each dataset.

\begin{table}[t]
\small
\centering
\begin{tabular}{l p{1cm} p{0.7cm} p{2.5cm}}
\toprule
Benchmark & Input & Output & Annotation note \\ \midrule
\textcolor{red}{nvBench} & prompt, data & code, vis & Auto-generated based on a text-to-SQL benchmark \\
\textcolor{red}{ChartDialogs} & prompt, data, vis & code, vis & Manually annotated \\ 
\textcolor{red}{PlotCoder} & prompt, code & code & Auto-extracted \\
\bottomrule
\end{tabular}
\caption{Description of benchmark datasets.}
\label{tab:benchmark-description}
\end{table}

\section{Cross-language Mapping Table}
\label{sec:cross-language-mapping-table}

\begin{figure}[t!]
\centering
\includegraphics[width=\columnwidth]{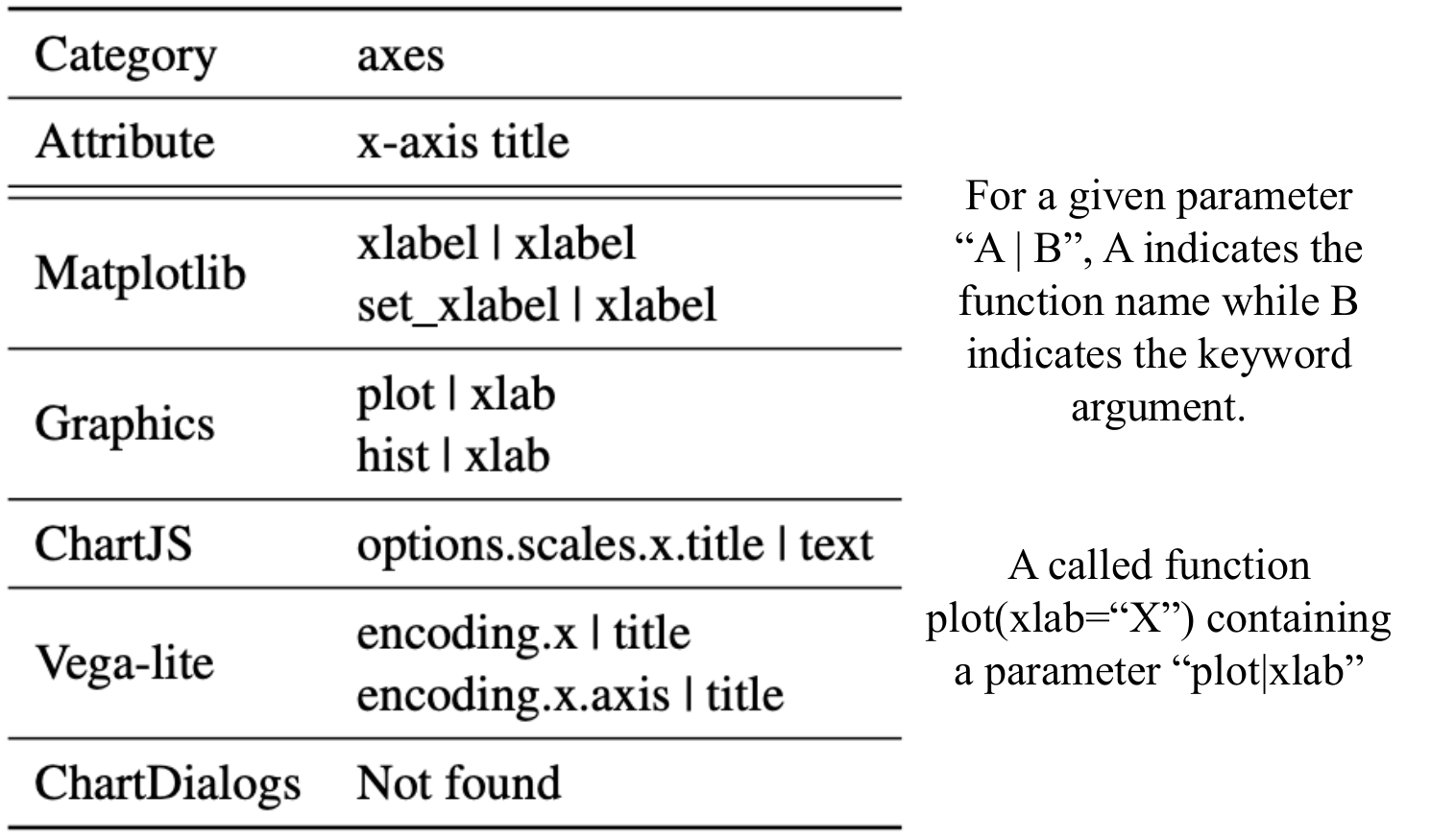}
\caption{Example of the cross-language mapping} 
\label{fig:sample-cross-lang-mapping-table}
\end{figure}

To compare the data described in the previous section, we constructed a cross-language mapping table based on frequently used parameters.
This involved selecting the top 500 frequently used parameters, identifying categories and attributes, and checking correctness based on the libraries' documentation and code execution.
Ultimately, the mapping table comprises 8 categories, 62 attributes, and around 850 parameters across the 4 visualisation languages. 
Figure~\ref{fig:sample-cross-lang-mapping-table} shows an example for the attribute ``x-axis title''.
Details can be found in Appendix~\ref{sec:appendix-cross-language-mapping-table}.

\section{Analysis \& Discussion}

\subsection{Comparison of chart types}
\label{ssec:common-plot-type}
\begin{figure}[t!]
\centering
\resizebox{1.05\columnwidth}{!}{
\includegraphics[]{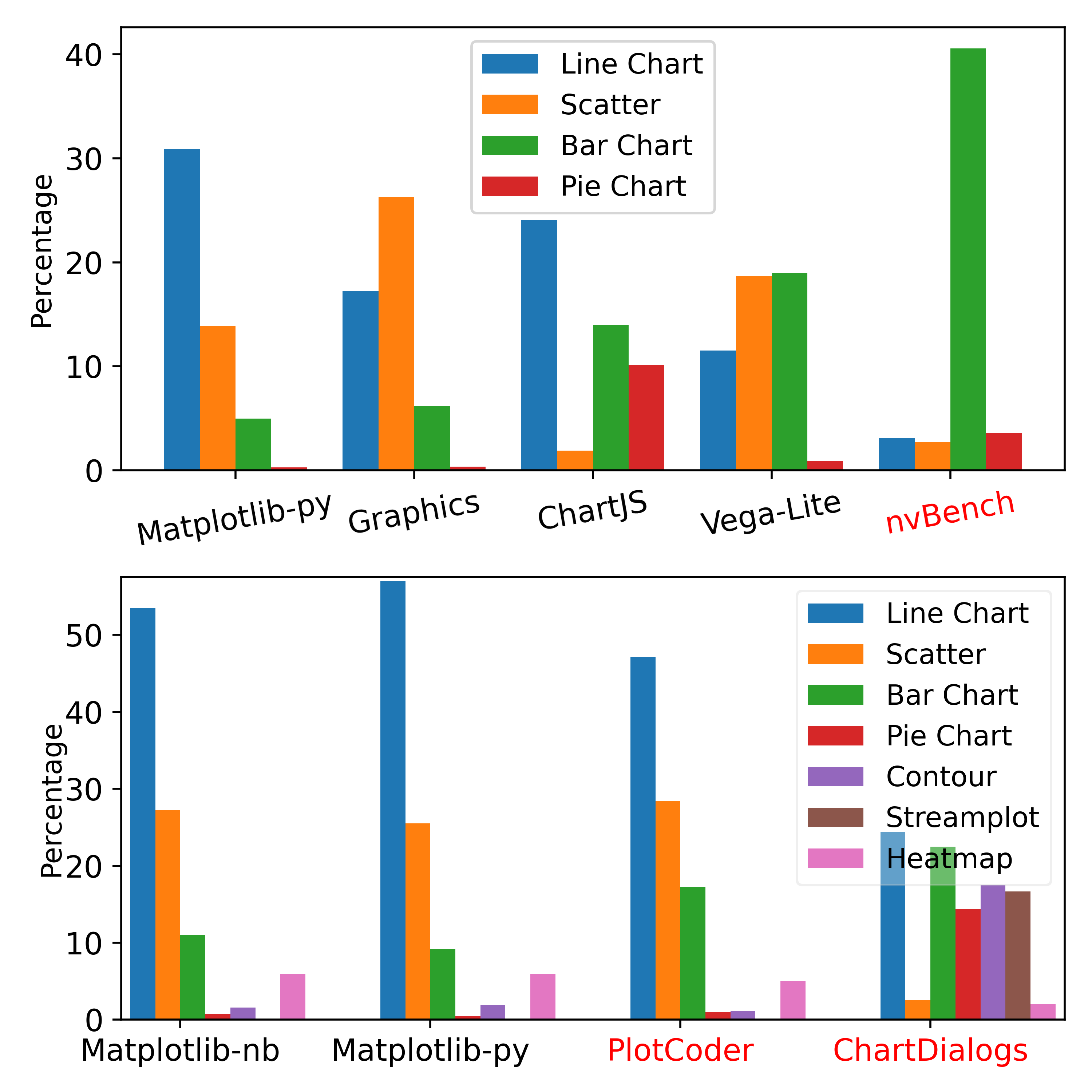}
}
\caption{Plot type distribution over eight datasets.} 
\label{fig:common-plot-type}
\end{figure}

Figure~\ref{fig:common-plot-type} (upper) depicts the distribution of four common plot types across real-world datasets and nvBench~\footnote{We categorise histograms as bar charts, while polar pie and doughnut charts are grouped as pie charts}. 
Each dataset shows distinct preferences for specific plot types.
The distribution of nvBench, a benchmark based on the Vega-Lite grammar, is significantly misaligned with that of Vega-Lite, where the bar chart dominates other types, accounting for over 80\%, while the remaining are around 7\%.

Figure~\ref{fig:common-plot-type} (lower) depicts the distribution of seven plot types across four Python-based datasets. 
Generally, the distribution between Matplotlib and PlotCoder shows notable similarity.
This trend is because both are derived from GitHub.
In contrast, ChartDialogs contains a more uniform distribution of plot types.
This is the result of its design, and differs from what we observe in the wild.
Specifically, ChartDialogs has fewer scatter plots and an overabundance of pie charts, contours, and stream plots.

These findings imply that nvBench and ChartDialogs are not testing the same distribution of plot types as real-world data.
As suggestions for future dataset makers, it is crucial to tailor the distribution of chart types according to the specific needs and domains of the intended users.
At the same time, given the imbalanced distributions in real-world data, it is also valuable to conduct separate evaluations focusing specifically on rarer plot types, acknowledging their distinct value.

\subsection{Comparison of attributes}
\label{ssec:common-visual-attributes}
\begin{figure}[t]
\centering
\includegraphics[width=0.95\columnwidth]{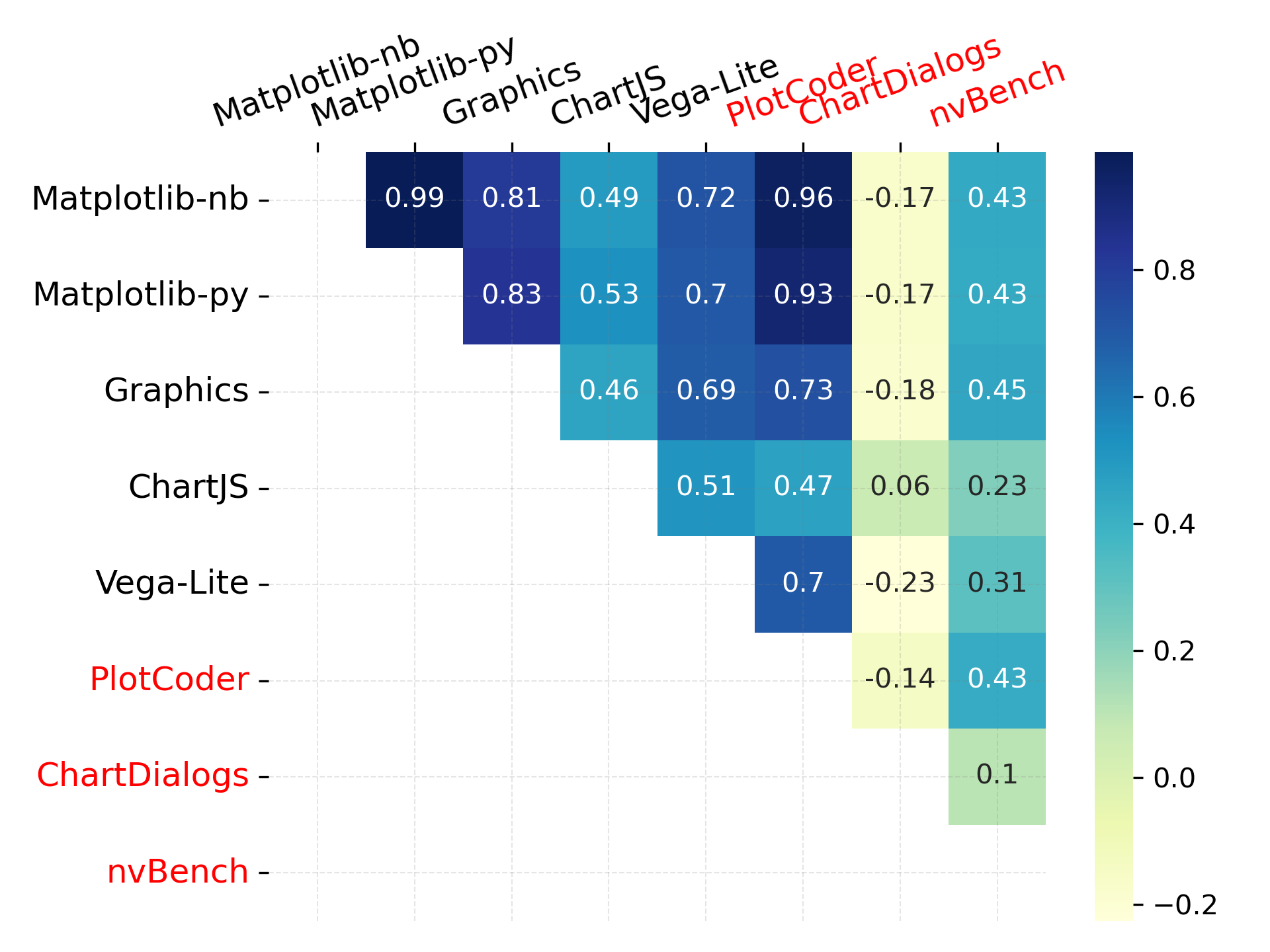}
\caption{Spearman's rank correlation coefficient in terms of frequent attributes} 
\label{fig:spearman-correlation-heatmap}
\end{figure}

Using the cross-language mapping table and parsed data (function names and arguments), we computed the normalised frequency for 62 attributes within each dataset, as shown in Figure~\ref{fig:common-visual-aspects} located in Appendix~\ref{sec:appendix-common-attributes}.
We used these frequencies to determine the Spearman’s rank correlation coefficient across eight datasets, as illustrated in Figure~\ref{fig:spearman-correlation-heatmap}.

The real-world datasets have a significant correlation, with Spearman's values surpassing 0.7, except for ChartJS, which displays a moderate correlation with coefficients around 0.5. 
As for the benchmarks, ChartDialogs and nvBench show a weak correlation with their direct counterparts, Matplotlib and Vega-Lite, respectively.
This means many attributes that were frequently used by end users have not been tested in these benchmarks.
These include titles, axes-scale limits, tick labels, opacity, histogram bins, legend visibility, and multiple plots handling, as visualised in Figure~\ref{fig:common-visual-aspects} in Appendix~\ref{sec:appendix-common-attributes}.

Conversely, PlotCoder demonstrates a strong alignment with real-world data, with Spearman's values ranging from 0.7 to 0.9. 
The correlation highlights PlotCoder's potential as a resource for crafting end-to-end Text-to-Vis benchmarks.
However, it is an automatically extracted dataset.
It lacks data to plot in the input, visualisations in the output, and information on which versions of libraries it has as dependencies.
This means the code cannot be executed as is.
Without executing it, there is not way to confirm whether the visual output aligns with user goals.

\subsection{Comparison of attributes when permitted}
\label{ssec:default-parameters}
Some attributes can only be activated (or permitted) if specific preconditions are met. 
For instance, the "bar thickness" attribute can only be set if the user plots data on a bar chart and adjusts the width parameter.
Consequently, these attributes may appear infrequently in the dataset, but users often specify their preferred values.
Therefore, analysing these attributes is crucial for a deeper understanding of end users' preferences.

In this analysis, we computed the frequency of attributes for a given visualisation type or action (e.g. \verb|plt.bar()|).
This calculation is applied to each attribute in the mapping table and visualised as a heat map in Figure~\ref{fig:default-parameters} located in Appendix~\ref{sec:appendix-common-attributes}.
We focus solely on examining this behaviour in Python-based datasets, including Matplotlib-nb, Matplotlib-py, PlotCoder, and ChartDialogs.
This is because nvBench does not prioritise user intention for modifying aesthetic attributes while others have different characteristics.

The Spearman's coefficient calculation among these datasets reinforces our findings in the previous section.
Matplotlib-nb, Matplotlib-py, and PlotCoder show significant correlations, with Spearman's scores above 0.8, whereas ChartDialogs scores below 0.1. 
While attributes such as axes' scales, edge colour, marker size, pie chart characteristics, legend labels, and grid line attributes receive considerable attention in ChartDialogs, end users less frequently specify them and often rely on the library's defaults.

Dataset creators should consider attributes such as histogram bins, pie precision digits, error-bar visibility, and annotation attributes, which are frequently customised by end users.

\subsection{Comparison of program complexity}
\label{ssec:number-of-functions-and-parameters}
To compare complexity, we calculate the average count of distinct visualisation functions and parameters within each code file and present the findings in Table~\ref{tab:number-of-funcs-and-params}. 
In this section, we omit ChartDialogs because it is a slot-filling dataset, with a fixed number of functions and parameters.

Benchmarks differ significantly from their direct counterparts.
They use far fewer functions and parameters.
In most real-world data, users employ 3 to 7 functions and 10 to 14 parameters.
The top 7 functions used in Matplotlib-py include plotting the data, saving figures, assigning titles, and adjusting legends. 
The higher figures in the Vega-Lite dataset can be explained by its nature as a visualisation language (not a library built on top of a programming language).

\begin{table}[t]
\centering
\small
\begin{tabular}{ l c c}
\toprule
            & No. Funcs &  No. Params \\ \midrule
py-Matplotlib  & 6.40          & 10.61          \\
nb-Matplotlib  & 6.19          & 10.54          \\ 
\textcolor{red}{PlotCoder}    & 4.05          & 6.03           \\ \midrule
Graphics    & 3.20          & 13.82          \\ \midrule
ChartJS     & 6.51          & 12.08          \\ \midrule
Vega-Lite   & 10.73         & 19.27          \\ 
\textcolor{red}{nvBench}     & 4.59          & 10.02          \\ \bottomrule
\end{tabular}
\caption{Average number of functions and parameters}
\label{tab:number-of-funcs-and-params}
\end{table}

\section{Conclusion}
In this paper, we analysed whether Text-to-Vis benchmarks accurately reflect real-world usage by presenting analyses of chart types, frequent attributes, and program complexity. Our results show that only one of the standard three benchmarks is aligned with real-world use. 
That dataset has its own critical limitation: it cannot be used as an end-to-end benchmark, going from a request and data as input to a visualisation as output. 
As well as critiquing current benchmarks, we provide guidance for future benchmark development, suggesting the evaluation of relevant attributes and challenging charts that better reflect end users' preferences. 
Such a benchmark would guide the development of useful and impactful systems.

\section*{Limitations}
This study offers analyses of datasets and acknowledges several limitations.
Firstly, our examination was restricted to only four visualisation libraries, each corresponding to a different programming language. 
This narrow scope may not adequately capture the diversity of applications and use cases within the field. 
Although we attempted to analyse MatLab code files in The Stack dataset, they are miscategorised in The Stack, processed with the wrong extension.\footnote{https://huggingface.co/datasets/bigcode/the-stack-dedup/blob/main/programming-languages.json}
Despite our efforts to clarify this issue by reaching out to the project authors, we have yet to receive a response.
Secondly, our investigation is based on public code, mainly representing programmers with different visualisation levels, including novices, practitioners, and experts.
If the target users are in a visualisation application like Tableau~\footnote{https://www.tableau.com/}, our results may not be representative.
Lastly, this study concludes with an analysis and assessment of existing benchmark datasets without proposing solutions. 
Nevertheless, we believe that the insights and recommendations provided in this work are valuable for any dataset maker and future studies.

\section*{Ethics Statement}
The data used in this research can be found publicly in the repositories of the cited papers, GitHub, or HuggingFace.
Those who want to use the processed data in our repository will need to follow the terms and conditions of The Stack dataset\footnote{https://huggingface.co/datasets/bigcode/the-stack-dedup}.

\section*{Acknowledgments}
This material is partially supported by the Australian Research Council through a Discovery Early Career Researcher Award and the Commonwealth Scientific and Industrial Research Organisation (CSIRO). We extend our gratitude to the anonymous reviewers for their constructive feedback and valuable advice on our submissions.

\bibliography{acl2023}
\bibliographystyle{acl_natbib}

\appendix

\section{Code Parsing}
\label{sec:appendix-code-parsing}

After obtaining code files for Python and R, we used abstract syntax tree (AST) parsers and heuristics to accurately extract variables, function names, arguments, and explicit values.
Subsequently, we tracked the assigned variables to correctly select the functions used in Matplotlib while a list of Graphics' functions was used to filter for this library.

To extract ChartJS specifications, we initially used an AST parser to extract all JSON data from the Javascript code files. 
Subsequently, a heuristic selection method was applied to filter JSON containing the three essential components of this library, namely "type," "data," and "options." 
This is because ChartJS relies on the JSON format as its foundation, serving as the input for executing functions in Javascript. 

Vega-Lite can appear in both JSON and Javascript files, as it is a JSON schema visualisation language. Therefore, we used the above methods for extraction.
In detail, after extracting JSON data from code files, we exclusively extracted snippets containing Vega-Lite schema ~\footnote{Vega-Lite schema: \href{https://vega.github.io/schema/vega-lite/v1.json}{v1},
\href{https://vega.github.io/schema/vega-lite/v2.json}{v2},
\href{https://vega.github.io/schema/vega-lite/v3.json}{v3},
\href{https://vega.github.io/schema/vega-lite/v4.json}{v4},
\href{https://vega.github.io/schema/vega-lite/v5.json}{v5},}, which is a mandatory field of Vega-Lite specification.

\begin{figure}[t!]
\centering
\includegraphics[scale=0.3]{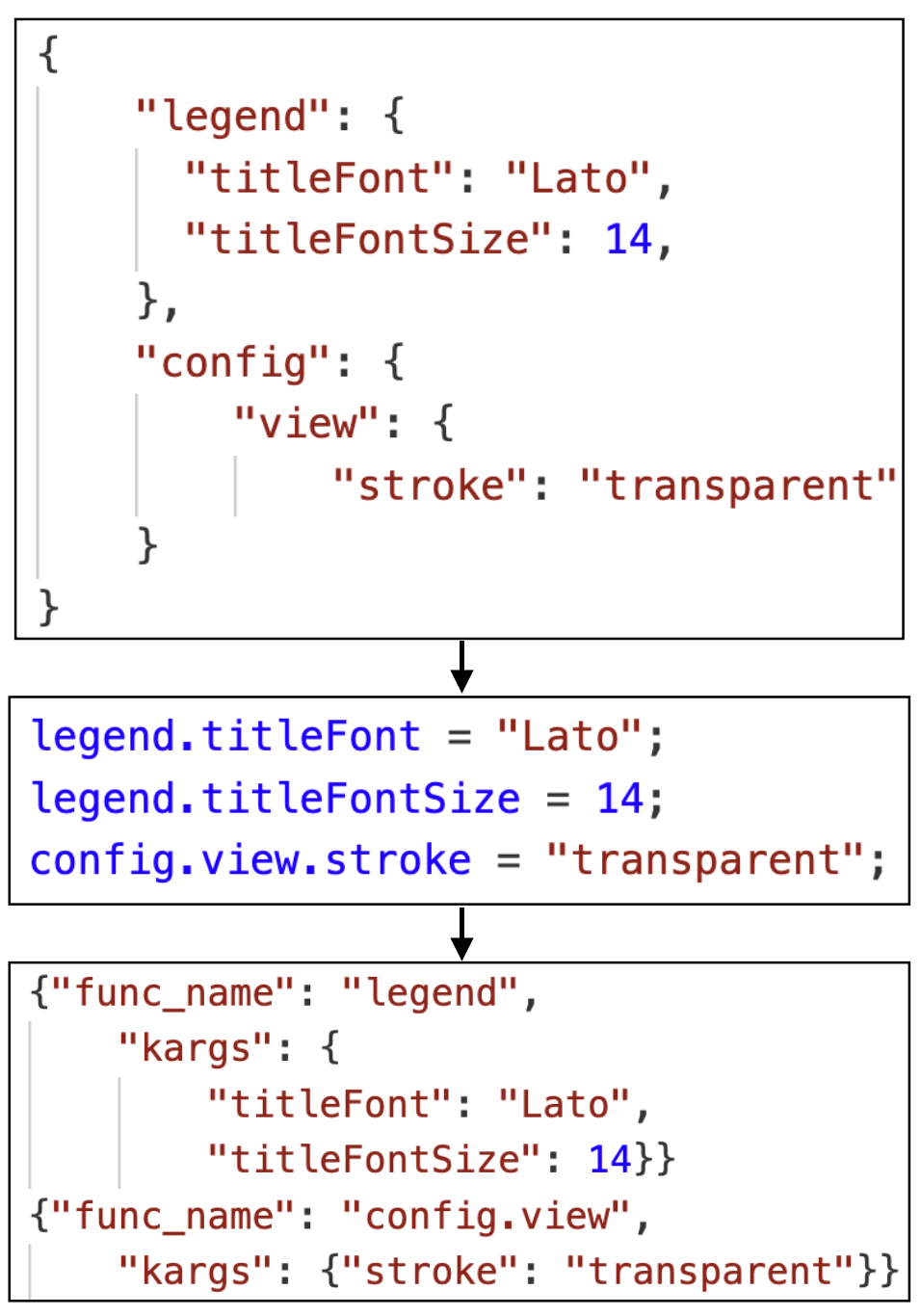}
\caption{The process of converting JSON to universal format} 
\label{fig:json-converting-process}
\end{figure}

After extracting functions, arguments, assigned values, and JSON specifications, targeting the visualisation libraries, we transformed them into a universal format to facilitate more accessible analysis and further processing.
For instance, a command in Python \verb|ax.plot(x, color='green', marker='o')|, which plots a line graph of `\verb|x|', with marker `\verb|o|' and colour `\verb|green|', can be parsed into a JSON as \verb|{"func_name": "plot", args: ["x"],|
\verb|kargs: {"color": "green", "marker": "o"}}|.
An example of translating JSON to universal format can be seen in Figure~\ref{fig:json-converting-process}.

Regarding \textcolor{red}{nvBench} and \textcolor{red}{PlotCoder}, they contain visualisation code in Vega-Lite and Python, so the process was the same as described above.
When it comes to \textcolor{red}{ChartDialogs}, a slot-filling dataset, we converted each user's intent to a function with changed slots as keyword parameters.
For example, a user's intent "smaller radius, increase text size" modifying a pie chart is transformed as universal JSON format \verb|{'func_name':'pie', kargs:{'radius:|
\verb|'small', 'font_size':'large'}}|.

\section{Benchmarks' Examples}
\label{sec:appendix_benchmark_examples}
\begin{figure}[h!]
\centering
\includegraphics[width=0.8\columnwidth]{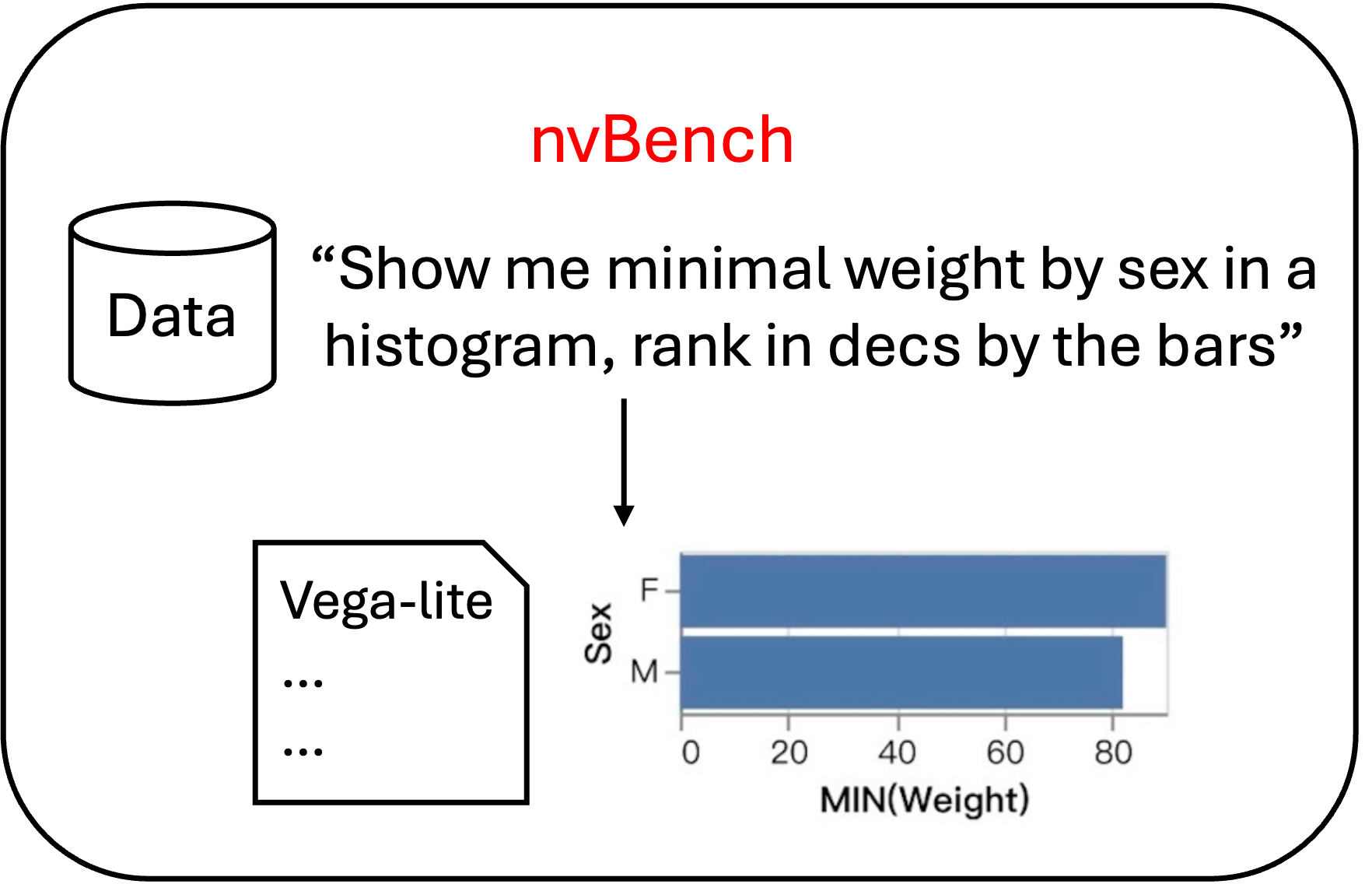}
\caption{A sample from the nvBench dataset.} 
\label{fig:nvbench_example}
\end{figure}

\begin{figure}[h!]
\centering
\includegraphics[width=0.8\columnwidth]{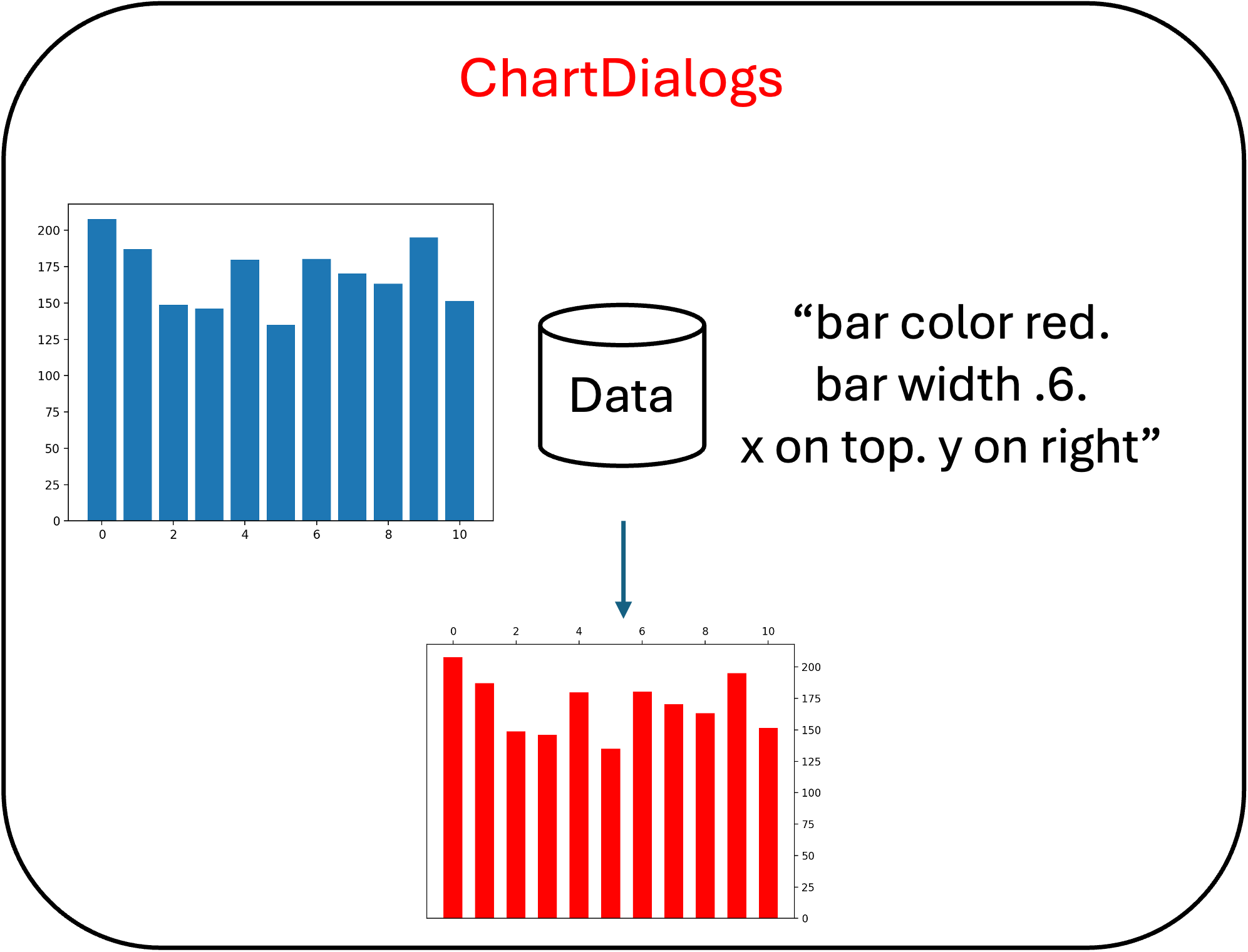}
\caption{A sample in ChartDialogs dataset. This dataset was built in a slot-filling manner. The visualisation is generated by a hard-coded program.} 
\label{fig:chartdialog_example}
\end{figure}

\begin{figure}[h!]
\centering
\includegraphics[width=0.8\columnwidth]{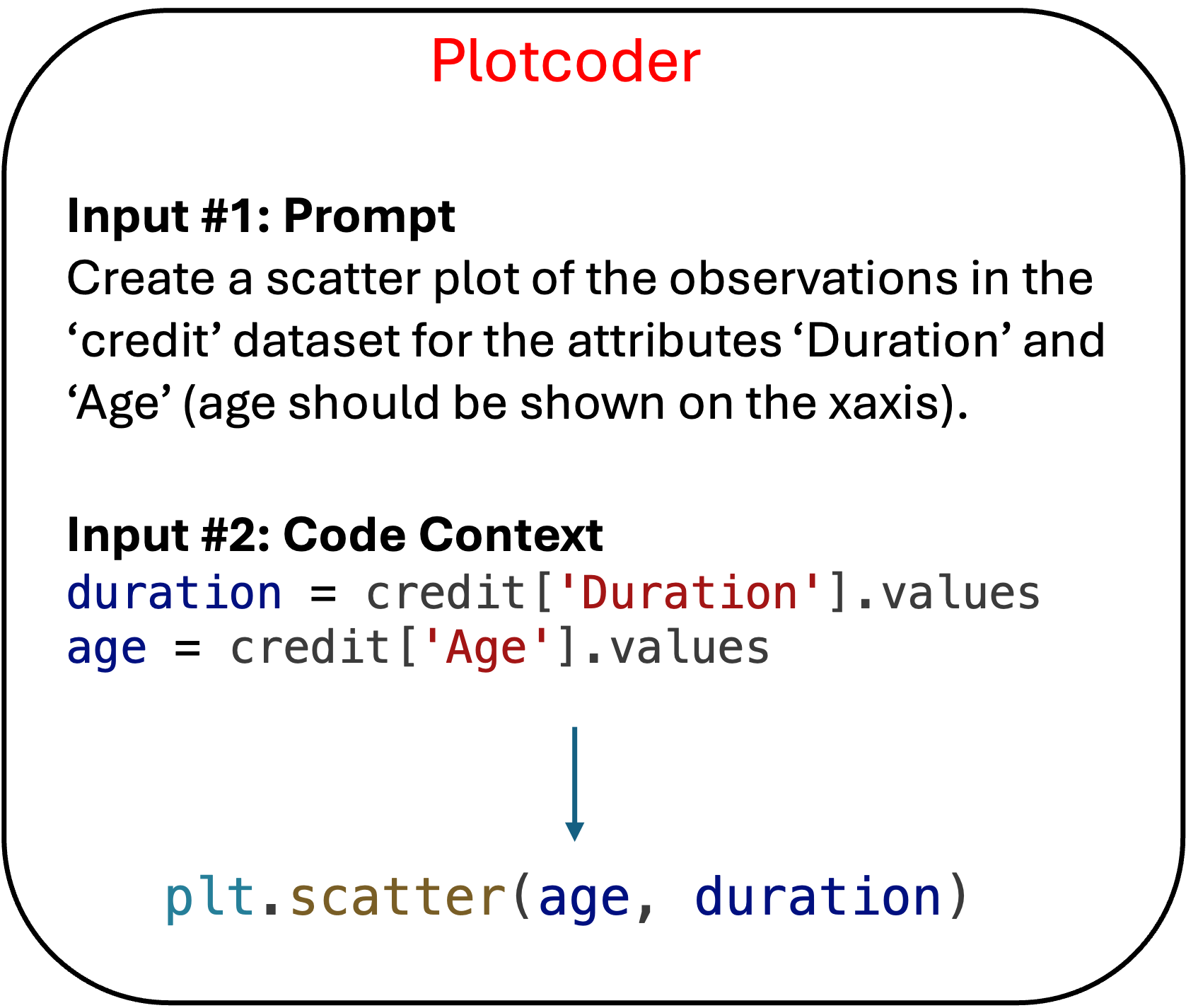}
\caption{A sample from the PlotCoder dataset.} 
\label{fig:plotcoder_example}
\end{figure}

Figures~\ref{fig:nvbench_example},~\ref{fig:chartdialog_example}, and~\ref{fig:plotcoder_example} illustrate examples for nvBench, ChartDialogs, and PlotCoder, respectively.

\section{Cross-language Mapping Table}
\label{sec:appendix-cross-language-mapping-table}
The procedure for making the cross-language table is as follows.
Initially, we compiled the top 100 frequent parameters from the real-world dataset in 4 languages: Matplotlib, Graphics, ChartJS, and Vega-Lite. 
Subsequently, the parameters were grouped into different categories and attributes.
The mapping table was further expanded by investigating relevant parameters within the top 500. 
If a specific language lacked relevant parameters for a given attribute in the top 500 (resulting in a blank cell), we persistently searched through the remaining list until a match was found.
Cells where no relevant parameter was identified led to the annotation of ``not found.''
This identification and verification process includes
understanding plotting parameters, identifying them in API documents, asking ChatGPT~\footnote{ \href{https://chat.openai.com}{https://chat.openai.com} 
\href{https://bard.google.com/chat}{https://bard.google.com/chat}
} for explanations and relevant parameters and executing example codes.

Table~\ref{tab:all-categories-and-attributes} shows a small part of the table for context.
The whole table can be found in our repository at \href{https://github.com/giahy2507/text-to-vis-bench-assessment}{https://github.com/giahy2507/text-to-vis-bench-assessment}.

\begin{table}[htp!]
\centering
\resizebox{\columnwidth}{!}{
\begin{tabular}{l l}
\hline
Category & Attribute \\ \hline
Axes & \begin{tabular}[c]{@{}l@{}}x-title, y-title, x-y-title-fontsize, x-y-title-color, \\ x-y-lim, x-y-ticks-labels, x-y-ticks-labels-color, \\ x-y-ticks-labels-rotation, x-y-scale, \\ x-y-ticks-fontsize, x-axis-ticks-visible, \\ y-axis-ticks-visible, x-y-scale-position, \\ invert-x-y-axis\end{tabular} \\ \hline
Data Appearance & \begin{tabular}[c]{@{}l@{}}filled-color, edge-color, opacity, linewidth, \\ markersize, linestyle, line-capstyle, markerstyle, \\ bar-thickness, bar-data-stacking, hist-bins, \\ pie-explode, pie-label-distance, \\ pie-percentage-distance, pie-precision-digits, \\ pie-radius, errbar-cap-size, errbar-cap-thick, \\ errbar-color, errbar-visible\end{tabular} \\ \hline
Annotation & \begin{tabular}[c]{@{}l@{}}ann-text/label, ann-fontsize, ann-possition, \\ ann-font\end{tabular} \\ \hline
Main title & \begin{tabular}[c]{@{}l@{}}title, title-fontsize, title-color, title-position, \\ subtitle, subtitle-fontsize\end{tabular} \\ \hline
Legend & \begin{tabular}[c]{@{}l@{}}legend-title, legend-fontsize, legend-position, \\ legend-labels, legend-labels-color, \\ legend-is-display\end{tabular} \\ \hline
Grid & \begin{tabular}[c]{@{}l@{}}grid-visible, grid-color, grid-linestyle, \\ grid-linewidth\end{tabular} \\ \hline
Format & size, dpi, saving-format \\ \hline
Other & \begin{tabular}[c]{@{}l@{}}bounding-box/border, background, \\ margin/padding, multiple-plots\end{tabular} \\ \hline
\end{tabular}
}
\caption{Categories and Attributes in the cross-language mapping table.} 
\label{tab:all-categories-and-attributes}
\end{table}

\section{Calculation for heat map figures}
\label{sec:appendix-common-attributes}

Figure~\ref{fig:common-visual-aspects} shows heat maps of the most common visualisation attributes over 7 datasets, where the more intense green colour indicates a higher percentage of usage within the dataset. 
The calculation for each attribute is $k/n$, where:
\begin{itemize}
    \item $k$ is the number of times that attribute's arguments are specified
    \item $n$ is the number of times that all arguments are specified
\end{itemize}

As for the heat map in Figure~\ref{fig:default-parameters}, there are two cases influencing different levels.
For attributes impacting the program level, such as title, x-axis title, and x-y tick labels, the percentage is derived from how frequently a program includes arguments for a specific attribute. 
Conversely, for local attributes affecting the function level, like filled colour, opacity, and bar thickness, the percentage is calculated based on the frequency of functions containing arguments for the given attribute. 
The calculation is as follows:
\begin{itemize}
    \item \textbf{$k$} is the number of times that attribute's arguments are specified. \textbf{$p$} is the number of times that attribute's functions are used
    \item $z$ represents the total number of programs in the dataset, while $g$ denotes the number of programs in which the attribute is used (any of the attribute arguments is used).
\end{itemize}
While a figure for a given program-level attribute is $g/z$, that for function-level one is $k/p$.

\begin{figure}[h!]
\centering
\resizebox{\columnwidth}{!}{

}
\caption{Heat map of the most frequent aesthetic attributes over 7 datasets. The attributes are classified by different categories with colours, such as \colorbox{xy-axis}{x and y axes}, \colorbox{data-appearance}{data appearance}, \colorbox{annotation}{annotation}, \colorbox{title-subtitle}{title and subtitle}, \colorbox{legend}{legend}, \colorbox{grid}{grid}, \colorbox{figure-format}{figure format}, and \colorbox{others}{others}.} 
\label{fig:common-visual-aspects}
\end{figure}

\begin{figure}[h!]
\centering
\resizebox{0.78\columnwidth}{!}{

}
\caption{Heat map of attributes that the user often specifies values when permitted.} 
\label{fig:default-parameters}
\end{figure}

\begin{table*}[h!]
\resizebox{\textwidth}{!}{
%
 & Not found \\ \bottomrule
\end{tabular}
}
\caption{Details of the cross-language mapping table for 7 attributes over 3 categories. Each parameter in attributes comprises two parts, function name and argument name, separated by ``$|$''.}
\label{fig:cross-lang-mapping-table-8-atts}
\end{table*}

\end{document}